\documentclass[twoside,11pt]{article}
\usepackage{jair, theapa, rawfonts, graphicx}

\jairheading{1}{Submitted}{}{11/21}{}
\ShortHeadings{A Framework for Fairness}
{Richardson \& Gilbert}
\firstpageno{1}

\begin{document}

\title{A Framework for Fairness: A Systematic Review of Existing Fair AI Solutions}

\author{\name Brianna Richardson \email richardsonb@ufl.edu \\
       \name Juan E. Gilbert \email juan@ufl.edu \\
       \addr University of Florida\\
       Gainesville, FL 32601 USA}


\maketitle

\begin{abstract}
In a world of daily emerging scientific inquisition and discovery, the prolific
launch of machine learning across industries comes to little surprise for those
familiar with the potential of ML. Neither so should the congruent expansion
of ethics-focused research that emerged as a response to issues of bias and unfairness that stemmed from those very same applications. Fairness research,
which focuses on techniques to combat algorithmic bias, is now more supported
than ever before. A large portion of fairness research has gone to producing
tools that machine learning practitioners can use to audit for bias while designing their algorithms. Nonetheless, there is a lack of application of these fairness
solutions in practice. This systematic review provides an in-depth summary of
the algorithmic bias issues that have been defined and the fairness solution space
that has been proposed. Moreover, this review provides an in-depth breakdown
of the caveats to the solution space that have arisen since their release and a
taxonomy of needs that have been proposed by machine learning practitioners,
fairness researchers, and institutional stakeholders. These needs have been organized and addressed to the parties most influential to their implementation,
which includes fairness researchers, organizations that produce ML algorithms,
and the machine learning practitioners themselves. These findings can be used
in the future to bridge the gap between practitioners and fairness experts and
inform the creation of usable fair ML toolkits.
\end{abstract}

\section{Introduction}
\label{Introduction}
Today, applications of machine learning (ML) and artificial intelligence (AI) can be found in nearly every domain \shortcite{Jordan2015}: from medicine and healthcare \shortcite{Goldenberg2019,Chang2018,Lin2020,Munavalli2020} to banking and finance \shortcite{Sunikka2011,Choi2018,Moysan2019}. At a rate that has grown exponentially over the last decade, companies are seizing the opportunity to automate and perfect procedures. In the field of healthcare, machine learning is being used to diagnose and treat prostate cancer \shortcite{Goldenberg2019}, perform robotic surgeries \shortcite{Chang2018}, organize and schedule patients \shortcite{Munavalli2020}, and digitize electronic health record data \shortcite{Lin2020}. Within banking and finance, machine learning is being used to personalize recommendations for consumers \shortcite{Sunikka2011}, detect and prevent instances of fraud \shortcite{Choi2018}, and provide faster and personalized services through the use of chatbots \shortcite{Moysan2019}. 

While machine learning is praised for its ability to speed up time-consuming processes, automate mundane procedures, and improve accuracy and performance of tasks, it is also praised for its ability to remain neutral and void of human bias. This is what \shortciteA{Sandvig2014} calls the ‘neutrality fallacy’, which is the common misconception that AI does not perpetuate the trends of its data which is often clouded with human biases. For this reason, machine learning continues to undergo heavy scrutiny for the role it plays in furthering social inequities. Over the past few years, major companies have been headlined for their AI technologies that cause harms against consumers. \shortciteA{Buolamwini2018} assessed commercially-used facial recognition systems and found that darker-skinned women were the most misclassified demographic group; \shortciteA{Noble2018} found that search engines perpetuated stereotypes and contributed substantially to representational harms; and \shortciteA{Angwin2016} found biases in automated recidivism scores given to criminal offenders. 

Circumstances such as these have led to a surplus of contributions to the solution space for algorithmic bias by stakeholders and researchers, alike \shortcite{Zhong2018}. Institutions have been quick to formulate their own ethics codes centered around concepts like fairness, transparency, accountability, and trust \shortcite{Jobin2019}. New conferences, like ACM’s Fairness, Accountability, and Transparency (FAccT) conference\footnote{ACM FAccT Conference - https://facctconference.org/}, and new workshops have emerged centered on these concepts. Furthermore, in most machine learning conferences, new tracks have been added that focus on algorithmic bias research. This surplus of recognition by stakeholders and computer science researchers has encouraged a commensurate rise in related contributions to improve the ethical concerns surrounding AI. 

One major objective for responsible AI researchers is the creation of fairness tools that ML practitioners across domains can use in their application of responsible and ethical AI. These tools translate top-tier research from the responsible AI space into actionable procedures or functions that practitioners can implement into their pipelines. Despite the number of currently existing tools, there is still a lack of application of ethical AI found in industry. Recent research suggests a disconnect between the fairness AI researchers creating these tools and the ML practitioners who are meant to apply them \shortcite{Holstein,Law,Madaio2020,Law2020,Veale}. \shortciteA{Greene2019} emphasizes that the solution to this problem lies in the intersection of technical and design expertise. Responsible and fair AI must undergo deliberate design procedures to match the needs of practitioners and satisfy the technical dilemmas found in bias research.

This survey paper will highlight users' expectations when engaging with fair AI tooling, and it will summarize both fairness concerns and design flaws discussed in literature. First, Section \ref{problem} focuses on algorithmic bias and the major entry points for bias that perpetuate the need for fairness research and fairness tools. Section \ref{solution} focuses on the solutions that have been posed thus far by fairness researchers. It will highlight major contributions and the features offered by them. Section \ref{recs} will focus on how these solutions are working in practice. It will highlight many of the drawbacks and provide recommendations for fairness experts, organizations, and ML practitioners. This paper is the first of its kind to provide a comprehensive review of the solution space of fair AI.

\section{The Problem Space: Algorithmic Bias}
\label{problem}

The first recorded case of algorithmic bias was the discrimination suit that was filed against St. George’s Hospital Medical School in the 1980’s \shortcite{Lowry1988}. Leaders of the admission program had decided to create an algorithm that could mimic the admission process by completing the first screening of applicants. Upon completion of the algorithm, they validated its performance by comparing its results to manually generated decisions and found a 90-95\% similarity. After a few years in circulation, staff began to notice trends in admission and brought their concerns to the attention of the school’s internal review board (IRB). When the IRB agreed that the correlation between machine scores and human scores delegitimized claims of bias, those claims were taken to the U.K. Commission for Racial Equity. After thorough analysis of the algorithm, the Commission confirmed these claims were true: the algorithm was placing value on applicant’s names and place of birth, penalizing individual’s with ‘Non-Caucasian’ sounding names \shortcite{Lowry1988}. 

The ruling of this committee was pivotal to the start of algorithmic bias research since it set a precedent that the inclusion of bias in a system, even for the sake of accuracy, was impermissible. Nonetheless, for the past 40 years, this issue has been recurring. Biases continue to emerge in the creation and use of machine learning, legitimizing unfair and biased practices across a multitude of industries \shortcite{Lum2016}. In the often-cited 2017 presentation at NeurIPs, \shortciteA{Crawford2017} discusses two potential harms from algorithmic bias: representational and allocative harms. Representational harms are problems that might arise from the troublesome ways certain populations are represented in the feature space, and allocative harms are problems that arise from how decisions are allocated to certain populations \shortcite{Crawford2013}. Literature has demonstrated that bias can arise in all shapes and forms, and the task for fairness experts is to organize and confront those biases.

\begin{figure}
    \centering
    \includegraphics[width=\textwidth]{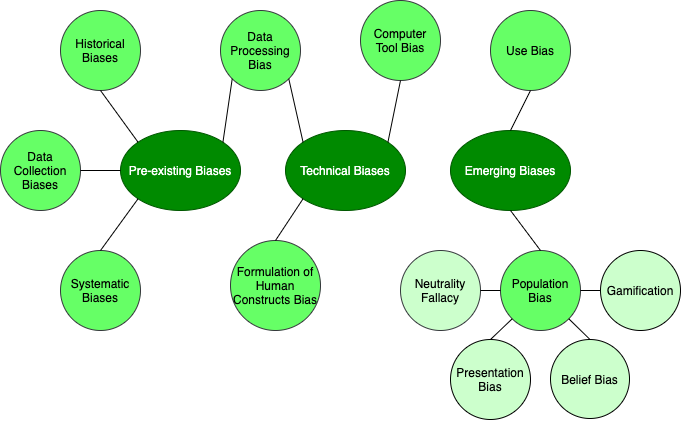}
    \caption{A taxonomy of the currently existing forms of bias.}
    \label{biases}
\end{figure}

\shortciteA{Friedman1996} separates types of biases into three categories: pre-existing, technical and emergent biases. They define pre-existing biases as those that are rooted in institutions, practices, and attitudes. Technical biases are those that stem from technical constraints or issues stemming from the technical design of the algorithm. Lastly, emergent biases arise in the context of real use by the user \shortcite{Friedman1996}. This section will adopt this categorization to classify influences of bias found in literature. A visual depiction of the forms of biases that will be explained can be seen in Figure \ref{biases}.

\subsubsection{Pre-existing bias} A large portion of literature attention has gone to pre-existing biases. These are biases that are associated with an individual or institutions. When human preferences or societal stereotypes influence the data and/or the model, that is said to be the effect of pre-existing bias. The saying, ‘garbage in, garbage out’ is well-known in the data science community as a euphemism for the quality of data you give a system will be the quality of your results. Fairness researchers translate this to mean that machine learning models serve as a feedback loop, reflecting the biases it has been fed \shortcite{Barocas2019}. Bias can come from an assortium of stages in the machine learning pipeline. 

The most prominently discussed form of bias is historical bias, or biases that are perpetuated in data and are the result of issues that existed at the time \shortcite{Veale,Calders2013,Olteanu2019,Barocas2019,Rovatsos2019,Suresh2020,Guszcza2018,Hellstrom2020}. \shortciteA{Suresh2020} defines historical biases as those that arise out of the misalignment between the world and its values and what the model perpetuates. The impact of historical bias can be reflected in what data was collected, how the data was collected, the quality of the data, and even the labels given to the samples \shortcite{Suresh2020,Rovatsos2019,Barocas2018,Calders2013,Hellstrom2020}. The subjectivity of labels is a major concern because it reproduces and normalizes the intentional or unintentional biases of the original labeler \shortcite{Barocas2018,Kasy2021}. From analyzing  a large collection of ML publications, \shortciteA{Geiger2019} found a dire need for more normalized and rigorous standards for evaluating datasets and data collection strategies prior to any machine learning processing, aligning with the recommendations from \shortciteA{Crawford2013}. 

Besides the historical biases, there also exists data collection biases, or biases that stem from the selection methods used for data sources \shortcite{Olteanu2019}. Data collection bias umbrellas a variety of different biases recognized in literature, including: sampling bias and representation bias. Sampling bias consists of issues that arise from the non-random sampling of subgroups that might lead to the under- or over-sampling of certain populations \shortcite{Mehrabi,Hellstrom2020}. Historically, vulnerable populations have often been undersampled \shortcite{Veale}. \shortciteA{Mester2017} discusses two types of sampling bias: selection bias and self-selection bias. While selection bias focuses on the biases of the data collector, self-selection bias stems from which individuals were available and willing to participate in the data collection process \shortcite{Mester2017}. Concerns about sampling bias stem from the fact that inferences made from mis-balanced dataset will inevitably lead to the disadvantage of populations who were missampled \shortcite{Asaro2019}. 

Representation bias, also called data bias \shortcite{Olteanu2019}, is another form of data collection bias and consists of bias that arises from the insufficient representation of subgroups within the dataset \shortcite{Crawford2017,Suresh2020}. This mis-representation can stem from the subjective selection of unfitting attributes or the incomplete collection of data \shortcite{Calders2013}. \shortciteA{Veale} describes this issue in detail, stating that subgroups can contain nuanced patterns that are difficult (or impossible) to quantify in a dataset, resulting in models that misrepresent those populations.

Another prominent form of pre-existing bias is data processing bias, or bias that is introduced by data processing procedures \shortcite{Olteanu2019}. The selection of methods across the ML pipeline can be incredibly subjective and many of these steps can introduce unintended biases into the model. For example, the use of sensitive attributes within models is a well-known taboo in data science. Government regulation has even gone as far as prohibiting organizations access to protected data \shortcite{Veale}. Nonetheless, the use of proxies, or attributes that encode sensitive information, is prolific \shortcite{Calders2013,Corbett-Davies2018}. Besides the use of sensitive data, other processing steps could incorporate bias, including the categorization of data \shortcite{Veale}, feature selection and feature engineering \shortcite{Veale,Barocas2018}, or even how data is evaluated \shortcite{Suresh2020}. Evaluation bias can occur from the use of performance metrics that are ill-fitting for the given model \shortcite{Suresh2020}.

The final category of pre-existing bias is systematic bias, or bias that stems from institutional or governmental regulations and procedures. While regulations limiting access to sensitive data can be beneficial for limiting use of sensitive data, they can also prevent the identification and de-biasing of proxy attributes leading to problematic models \shortcite{Veale}. Funding can also produce bias by manipulating the priorities of institutions and, therefore, practitioners \shortcite{Mester2017}. 

\subsubsection{Technical bias} While pre-existing biases consider what the model is adopting from the data, technical biases consider the limitations of computers and how technology insufficiencies might create bias \shortcite{Friedman1996}. \shortciteA{Baeza-Yates2018} considers this ‘algorithmic bias’ and defines it as bias that was not present in the data but was added by the algorithm. Overlapping on pre-existing bias, technical bias can also stem from processing procedures. Often, the selection of features, models, or training procedures can introduce bias that is not affiliated with the practitioner but with the insufficiency of those procedures to characterize the data. For example, hyperparameter tuning may, in an attempt to reduce the sparsity of the model, end up removing distinct patterns found in sub-populations \shortcite{Veale,Hellstrom2020}. Furthermore, certain models, such as regression, fail to capture the correlation between attributes and subgroups \shortcite{Veale,Skitka1999}. 

Computer tools bias is another form of technical bias that originates from the limitations of statistics and technology \shortcite{Friedman1996}. Simpson’s paradox is a statistical issue that arises from the aggregation of distinct subgroup data that produces misperformance for one or all subgroups \shortcite{Blyth1972,Suresh2020}.

Lastly, \shortciteA{Friedman1996} defines formulation of human constructs bias as that which stems from the inability of technology to properly grasp human constructs. In situations such as this, fairness experts often insist that practitioners or organizations reconsider the use of machine learning \shortcite{Rakova2020}. \shortciteA{Selbst2019} describes the effects of this bias as the ripple effect trap, where they describe how the use of technology in areas where human constructs bias exists can potentially change human values. 

\subsubsection{Emerging bias} This final category of bias, referred to by \shortciteA{Suresh2020} as deployment bias, is defined as biases that result from the deployment of a model. Emerging bias can emerge in two different forms, via population bias and use bias. Population bias stems from the insufficiency of a model to represent its population or society after deployment \shortcite{Olteanu2019}. This can stem from new knowledge that has emerged and made a model irrelevant \shortcite{Friedman1996} or a mismatch between the sample population used to train the model and the population who is impacted by its deployment \shortcite{Friedman1996,Calders2013,Rovatsos2019}. \shortciteA{Selbst2019} refers to this failure to consider how a model works in context as the “portability trap”.

\textit{Use bias} umbrellas a variety of different ways in which the use of models can create biases. In a study conducted by \shortciteA{Skitka1999}, participants interacted with an autonomous aid which provided recommendations for a simulated flight task. The results depicted trends of commission where a user agreed with a recommendation even when it contradicted their training \shortcite{Skitka1999}. The neutrality fallacy introduced by \shortciteA{Sandvig2014} is a major pitfall to AI application because it forces users to believe the tool is correct, even when it should be obvious that it is not \shortcite{Sandvig2014}. \shortciteA{Skitka1999} suggests that users would be less willing to challenge a decision if that decision was made with an algorithm.

A related form of use bias is presentation bias, or bias that stems from how a model is deployed \shortcite{Baeza-Yates2018}. In these situations, users make assumptions about the presentation format of algorithms and may come to incorrect conclusions. For example, in a search algorithm, many users attribute the rank of the result to the relevancy, which may not be accurate in all cases \shortcite{Baeza-Yates2018}.

Belief bias is also a form of use bias that occurs when someone is so sure of themselves they ignore the results \shortcite{Mester2017}. This can also impact when they choose to use the results of a model and when they choose to ignore them \shortcite{Veale2018}, particularly when the model’s decision counteracts their own beliefs. \shortciteA{Veale2018} discusses decision-support design that can be used to prevent this selectional belief bias. 

Gamification is another use bias in which a person learns how to manipulate an algorithm \shortcite{Veale2018}. \shortciteA{Ciampaglia2017} introduces the concept of popularity bias, or preferences that strongly correlate to popularity, and discusses how these forms of bias are often gamified on the internet by bots and fake feedback.

The final case of use bias is the curse of knowledge bias: when you assume someone has the background you do and can use the model appropriately \shortcite{Mester2017}. This concern appears in various literature surrounding predictive policing technologies \shortcite{Asaro2019,BennettMoses2018,Ridgeway2013} where critics voice concerns about whether police understand the limitations of these technologies and are using them appropriately.

\shortciteA{Guszcza2018} details the importance of understanding the environment where these machine learning tools will be used and creating with those environments in mind.

\section{The Solution Space: Fairness Technologies}
\label{solution}
The solution space to algorithmic bias consists of a diverse array of solutions that contribute to the responsible AI space, including explainability, transparency, interpretability, and accountability \shortcite{Cheng2021}. While each solution has diverged from a different ethical dilemma that emerged from the machine learning production pipeline, the objective remains the same: to produce fair AI, or AI that is free from unintentional algorithmic bias. Fairness, as defined by \shortciteA{Mehrabi}, is \begin{quote}
    ``[T]he absence of any prejudice or favoritism toward
an individual or a group based on their inherent or acquired characteristics."
\end{quote} 
Fair AI, for the purpose of this paper, consists of solutions to combat algorithmic bias, which is often inclusive of top-tier solutions from explainability, transparency, interpretability, and accountability research.

Since the problem space of algorithmic bias is so large and diverse, a concerted effort has gone into making fairness tools that practitioners can use to employ state-of-the art fairness techniques within their ML pipelines. These solutions mainly come in two forms: software toolkits and checklists. Toolkits serve as functions accessible via programming languages that can be used to detect or mitigate biases. Checklists are extensive guides created by fairness experts that ML/AI practitioners can use to ensure the inclusion of ethical thought throughout their pipelines. While this section won’t provide a complete description of the solution space, it will highlight the diversity of tools provided by both organizations and academic institutions.

\subsection{Software Toolkits}
A variety of software toolkits currently exist, each with overlapping and distinct characteristics to identify them. While some are accessible via website portals \shortcite{Saleiro2018}, many exist as installable packages that can be imported into Python programs \shortcite{Bird2020,Srinivasan2020,Johnson2020,Wexler2019} or other languages used by ML practitioners \shortcite{Bellamy2018,Vasudevan2020}. The following selection of toolkits provides a diverse look of available solutions.

\subsubsection{Google’s Toolkits} Google provides two relevant toolkits: Fairness Indicators \footnote{Tensorflow's Fairness Evaluation and Visualization Toolkit - https://github.com/tensorflow/fairness-indicators} and the What-If toolkit (WIT) \shortcite{Wexler2019}. Both toolkits function as interactive widgets where users can provide their model(s), their performance and fairness metrics of choice, and the attributes on which slicing and evaluating will take place \shortcite{Richardson2021}. These toolkits are embedded within the Tensorflow package in Python, but users can build their own custom prediction functions if their model exists outside of Tensorflow. Unlike other toolkits, Google’s toolkits require that users provide a model \shortcite{Seng}. Additionally, both toolkits allow model comparison between, at most, 2 models. While Fairness Indicators works with binary and multiclass problems and allows intersectional analysis, WIT works with binary or regression problems. Within Fairness Indicators, users can create their own visualizations to compare models across different performance and fairness metrics. Users can select and deselect performance metrics to focus on. The WIT includes a features overview, a performance chart, and a data point editor. The features overview provides visualizations to depict the distribution of each feature with some summary statistics. The performance chart provides simulated results depicting the outcome of select fairness transformations, including an option to customize a fairness solution. Lastly, the data point editor provides custom visualizations of data points from the data and allows users to compare counterfactual points, make modifications to points to see how the results change, and plot partial dependence plots to determine model’s sensitivity to a feature.

\subsubsection{UChicago’s Aequitas} Aequitas is an audit report toolkit that can be accessed via command line, as a Python package, and a web application \shortcite{Saleiro2018}. Aequitas is a tool built for classification models and allows for the comparison of models. While Aequitas does not provide fairness mitigation techniques, it does evaluate models based on commonly used fairness criteria. Users can label sensitive attributes and Aequitas will calculate group metrics between sensitive subgroups and provide disparity scores that reflect the degree of unfairness. Aequitas also produces plots that record these fairness disparities and group metrics. Within these visualizations, users can choose to have groups colored based on whether or not subgroup disparities pass a set threshold for “fair” or “unfair” \shortcite{Saleiro2018}. This tool was built with two types of users in mind: data scientists/AI researchers who are building these AI tools and policymakers who are approving their use in practice \shortcite{Saleiro2018}. 

\subsubsection{IBM’s AI Fairness 360} AI Fairness 360 (AIF360) is a toolkit that provides both fairness detection and mitigation strategies \shortcite{Bellamy2018}. It can be used in both Python and R. The fairness metrics provided include general performance metrics, group fairness metrics, and individual fairness metrics. Users also have an assortment of mitigation strategies that they can apply to their models. These mitigation strategies can be applied across a variety of stages in the pipeline, including: pre-processing, while processing, and post-processing. In order to apply these strategies, users can distinguish between privileged and unprivileged subgroups on which to complete analysis. To assist users with learning, their website includes a wide selection of tutorials and web demos \shortcite{Bellamy2018}. While this toolkit does not provide its own visualization functions, it does provide guidance for how its functions can be used in conjunction with the Local Interpretable Model-Agnostic Explanations toolkit, aka LIME \shortcite{Ribeiro}. Similar to Aequitas, AIF360 was built with business users and ML developers, in mind \shortcite{Bellamy2018}. 

\subsubsection{LinkedIn’s Fairness Toolkit (LiFT)} LiFT is a recently released toolkit that provides fairness detection strategies for measuring fairness across a variety of metrics \shortcite{Vasudevan2020}. They provide an assortment of metrics, segregated based on whether they are for the data or the model outputs. Similar to AI Fairness 360, they do not produce their own visualizations. Unlike previous toolkits, they are built to be applied in Scala/Spark programs. Furthermore, this toolkit is unique in its flexibility and scalability, overcoming previous issues with measuring fairness in large datasets \shortcite{Vasudevan2020}.

\subsubsection{Other Solutions} An assortment of other toolkits exist, including Microsoft’s Fairlearn \shortcite{Bird2020}, ML Fairness Gym \shortcite{Srinivasan2020},  Scikit’s fairness tool \shortcite{Johnson2020}, PyMetrics Audit-AI \footnote{PyMetric's Audit-AI - https://github.com/pymetrics/audit-ai}, and new ones arise often. While these toolkits may differ in the fairness metrics they consider and the fairness mitigation strategies they may employ, the highlights of their features can be seen in the toolkits above: some are interactive, some provide visual support, some allow intersectional analysis, some focus on detection or mitigation alone, etc.

\subsection{Checklist}
Unlike toolkits, checklists mostly exist as documentation meant to guide readers through incorporating ethical and responsible AI throughout the lifecycle of their projects. While some are specifically meant for data scientists or machine learning engineers building the tools, other checklists are generally written with all relevant parties in mind. While toolkits provide more statistical support when it comes to strategies to detect and mitigate, checklists engage developers with questions and tasks to effectively ensure that ethical thought occurs from the idea formulation to the auditing after deployment \shortcite{Patil2018}.

\subsubsection{Deon} Deon is a fairness checklist created by DrivenData, a data scientist-led company that utilizes crowdsourcing and cutting-edge technology to tackle predictive problems with societal impact \footnote{Driven Data's Deon tool - https://deon.drivendata.org/}. The checklist is described as a “default” toolkit, but customizations are made for practitioners with domain-specific concerns. Deon is split into five areas: data collection, data storage, analysis, modeling, and deployment. To assist users with utilizing the checklist, each checklist item is accompanied with use cases to exemplify relevant concerns. One of Deon’s most unique features is its command line interface that practitioners can use to interact with the checklist.

\subsubsection{Microsoft’s AI Fairness Checklist} \shortciteA{Madaio2020} produced the AI Fairness Checklist. Unlike most checklists, this checklist was co-designed with the iterative feedback of practitioners.  The checklist is split into 6 different parts: Envision, Define, Prototype, Build, Launch, and Evolve. Envision, Define, and Prototype provide ethical considerations that would fit best in the initial planning and designing stages of a project; Build provides guidance for when AI is in the creation phases; and Launch and Evolve provide guidance for the deployment stages of the product. It also comes with a comprehensive Preamble that introduces the complexity of fairness, provides instructions for how the checklist should be used, and encourages practitioners and teams to personalize and customize the checklist to best fit their environment \shortcite{Madaio2020}.

\subsubsection{Legal and Ethics Checklist} \shortciteA{Lifshitz2020} provides a unique ethics checklist that focuses on legal considerations that should be noted in the creation of AI. Unlike related checklists, this checklist is not sectioned by stages in the AI lifecycle, but by legal priorities, including: human agency \& oversight, security \& safety, privacy \& data governance, transparency, accessibility, etc. This checklist provides a unique viewpoint from a lawyer for institutions and practitioners to use to avoid running into legal issues \shortcite{Lifshitz2020}.

\subsubsection{IBM’s AI FactSheets} AI FactSheets is a unique guide that provides a methodology for incorporating responsibility and transparency into the AI pipeline via documentation. While AI FactSheets is not described as a checklist, it does provide guidance for how to create documentation that depicts responsible AI practices. The creation of a FactSheet template is very domain-dependent, and \shortciteA{Arnold2019} provides a detailed methodology for how a FactSheet can be customized to the project. In satisfying the components of the FactSheets, institutions and teams can engender trust with consumers by increasing transparency and ensuring their products satisfy necessary ethical considerations \shortcite{Arnold2019}.

These checklists provide an overview of the differing features that can be found across the checklist landscape: some focus on legality, some are domain specific, and others have been built with the AI pipeline in mind. Other checklists and guides have been produced as well \shortcite{UnitedKingdomDepartmentofDigitalCultureMediaandSport2020,Manders-Huits2009,Gebru2018,Mitchell2019}, similarly structured and intended to assist practitioners with implementing fair or responsible practices in their pipelines.

\section{Shortcomings of Solutions \& Recommendations for the Future}
\label{recs}
While the solutions posed above have been a pivotal starting point for fairness research, much work has been done addressing the pitfalls of these contributions, as well. This section aims to discuss the prominent issues that have arisen in the literature and the corresponding recommendations that have been made for future work. While many of these recommendations were made to fairness researchers and designers, there also are recommendations for organizations and ML practitioners.

\subsection{Recommendations for Fairness Experts}
Towards the goal of universally ethical AI, much of the responsibility lies in the hands of fair AI researchers to define, design, and translate fairness into a palatable procedure that can be easily applied by practitioners and institutions. This requires that the process of producing fair AI undergo human-centered research and design procedure. Nonetheless, previous research has found that there exist major gaps between practitioners and fairness experts, indicating a lack of communication between those producing fair AI and those intended to apply it \shortcite{Holstein,Law,Madaio2020,Law2020,Veale}. The results from these works have been thematically categorized, accompanied by suggestions for improvement.

\subsubsection{Conflicting Fairness Metrics} Along with identifying 23 different types of biases and 6 different categories for discrimination, \shortciteA{Mehrabi} identified 10 different types of fairness metrics. Due to the vast quantity and similarity between these metrics, many authors have tried to categorize them into related bins. \shortciteA{Mehrabi} identified three different types of fairness metrics: individual, group, and subgroup. Individual fairness gives similar predictions to similar individuals, group fairness treats different groups equally, and subgroup fairness attempts to achieve a balance of both group and individual fairness \shortcite{Mehrabi}. 

\shortciteA{Barocas2019} also identified 19 proposed fairness metrics and also categorized them into three categories: independence, separation, and sufficiency. \shortciteA{Barocas2019} defines these categories through properties of the joint distribution of sensitive attribute A, the target variable Y, and the classifier or Score R. A random variable (A, R) satisfies independence if $A \perp R$; a random variable (R, A, Y) satisfies separation if $R \perp $ A $ \mid Y$; and a random variable (R, A, Y) satisfies sufficiency if $Y \perp $ A $ \mid R$ \shortcite{Barocas2019}. 

Lastly, in a comprehensive review and explanation of fairness metrics, \shortciteA{Verma2018} isolated 20 different fairness metrics and categorized them into 3 different categories: statistical measures, similarity measures, and causal reasoning. Statistical measures are those that depend on true positive, false positive, false negative, and true negative; similarity based measures attempt to address issues that are ignored by statistical measures by focusing on insensitive attributes as well; and causal reasoning uses causal graphs to draw relations between attributes to determine their influence on the outcome and allow users to understand where exactly bias is coming from and whether it is permissible \shortcite{Verma2018}.

While each author recognized legitimate patterns within and between metrics, these works highlight a major issue that exists: there are too many metrics for measuring unfairness with too few differences to delineate them \shortcite{Mehrabi}. Furthermore, major trade-offs exist between fairness metrics, making the selection of metrics highly situation-dependent \shortcite{Binns2018a}. \shortciteA{Verma2018} concludes that many of these metrics are advanced and require expert-level input, which in itself produces implicit biases. 

Furthermore, many works have identified conflicts between metrics that make them incompatible with each other, which forces individuals to choose \shortcite{Berk2021,Friedler2016,Kleinberg2017,Mittelstadt2019,Rovatsos2019}. Different metrics may emphasize different aspects of performance \shortcite{Japkowicz2011} and much work has been done comparing and contrasting these metrics \shortcite{Garcia-Gathright2018,Verma2018,Chouldechova2017,Corbett-Davies2018,Binns2018a,Fish2016,Kilbertus2017,Simoiu2017,Cramer2019,Zliobaite2015}. This decision is not one that should be taken lightly because the act of choosing a fairness metric is very political in that it valorizes one point of view while silencing another \shortcite{Bowker1999,Friedler2016}. According to \shortciteA{Rovatsos2019}, the task of choosing the right metric currently lies in the hands of practitioners, most of whom are unfamiliar with fairness research, and that is a heavy expectation considering the fact that society as a whole has not decided which ethical standards to prioritize. 

To support this issue, \shortciteA{Verma2018} suggests that more work is needed to clarify which definitions are appropriate for which situations. \shortciteA{Friedler2016} states that fairness experts must explicitly state the priorities of each fairness metric to ensure practitioners are making informed choices. Furthermore, \shortciteA{Friedler2018} emphasizes that new measures of fairness should only be introduced if that metric behaves fundamentally differently from those already proposed. 

\subsubsection{Other Metric-specific Pitfalls} In addition to the conflicting nature of fairness metrics, there exist several other pitfalls that have been recognized by researchers. \shortciteA{Kilbertus2017} discussed the insufficiency of "observational criteria", which are the criteria often used as sensitive attributes within toolkits. They emphasize that such methodologies are unable to confirm that protected attributes have a direct causal influence on results and instead, they propose two new metrics of causal reasoning: proxy discrimination and unresolved discrimination \shortcite{Kilbertus2017}. Proxy discrimination can be potentially inferred from a causal graph when there exists a path from a protected attribute to a predicted attribute that includes a proxy variable, and unresolved discrimination can be inferred from a causal graph if the only path from a protected attribute to the predicted attribute includes a resolving variable \shortcite{Kilbertus2017}.

Furthermore, \shortciteA{Friedler2018} found that many fairness metrics lack robustness. By simply modifying dataset composition and changing train-test splits, the results of their study depicted that many fairness criteria lacked stability. They proposed that only measures that depict stability and success should be used to report fairness.

\shortciteA{Hoffmann2019} discusses the lack of intersectional analysis in fairness criteria. Many metrics provide analysis for sensitive attributes, but only one-dimensionally. \shortciteA{Hoffmann2019} emphasizes that proposed metrics should strategically identify multi-dimensional correlations between attributes and outcomes. 

\shortciteA{Wagstaff2012} emphasizes that the use of abstract metrics distract from the problems specific to datasets and applications. Furthermore, the impact of the metrics cannot be inferred from the scores, themselves. Furthermore, \shortciteA{Kasy2021} demonstrates that many of the standard metrics legitimize the focus on merit, deterring individuals from questioning the legitimacy of the status quo. 

Additionally, there is a concern about the assumptions made by fairness experts when producing fairness metrics. Many metrics and available fairness tooling depend on access to sensitive attributes, which many practitioners do not have \shortcite{Holstein,Law2020,Rovatsos2019}. Additionally, there are legal restrictions against some fairness definitions \shortcite{Xiang2019}. For those with (legal) access to sensitive data, some practitioners are concerned with how the public would scrutinize the use of sensitive attributes, even for the detection of bias \shortcite{Rovatsos2019}. Furthermore, when studying political philosophy and how it connects to fairness in machine learning, \shortciteA{Binns2018a} notes that risks of incomplete fairness analysis arise when users are forced to adhere to a static set of prescribed protected classes, instead of doing thorough analysis to identify discrimination. To combat this issue, \shortciteA{Law2020} suggests that coarse-grained demographic info be utilized in fairness tools and proposed to practitioners. Furthermore, fairness experts should utilize and promote fairness metrics that do not rely on sensitive attributes \shortcite{Holstein,Law2020,Rovatsos2019}.

\subsubsection{Oversimplification of Fairness} A major concern in literature is the emphasis on technical solutions to algorithmic bias, which is a socio-technical problem. Many authors emphasize the need to supplement statistical definitions with social practices \shortcite{Veale2018,Madaio2020,Verma2018,Fazelpour2020,Jacobs2021,Birhane2021}. \shortciteA{Madaio2020} called the sole use of technical solutions, “ethics washing,” and \shortciteA{Selbst2019} describes the failure to account for the fact that fairness cannot be solely achieved through mathematical formulation as the “formalism trap”.  \shortciteA{Fazelpour2020} details that the practice of ethics washing can lead to misguided strategies for mitigating bias, and \shortciteA{Harcourt2007} emphasizes that the perceived success of these technical solutions stalls pursuits to achieve actual fairness with the aid of social practices. \shortciteA{Birhane2021} calls for a fundamental shift towards considering the relational factors and impact of machine learning from a societal standpoint.

\shortciteA{Fazelpour2020} recommends that mathematical assessment be supplemented with social assessment tools: like a thorough analysis of data collection strategies and understanding the assumptions made by different ML algorithms. Furthermore, toolkits and checklists for fair evaluation should avoid solely providing mathematical solutions.

\subsubsection{Misguided fairness objectives} In addition to the issue of ethics washing, other concerns have arisen concerning the trajectory of fairness research. \shortciteA{Hoffmann2019} critiques fairness research for its focus on avoiding disadvantage instead of understanding where advantage stems from. In support of better understanding where biases exist in the data, \shortciteA{Veale} proposes that more energy should be placed in data exploration, through the use of unsupervised learning as a strategy to identify hidden patterns. Futhermore, many fairness metrics and strategies rely on assumptions about the world that may be problematic in themselves. For example, \shortciteA{Hu2020} discusses how the use of the social concept of sex can perpetuate problematic and harmful sex discrimination.

\subsubsection{Making Fair AI Applicable} Works, such as those done by \shortciteA{Holstein,Veale2018,Rakova2020,Richardson2021}, are novel in their inclusion of practitioner feedback into fairness literature. A common theme that emerged from these papers was the lack of applicability that most practitioners held toward fairness tools and support. Practitioners interviewed by \shortciteA{Holstein} emphasized the need for domain-specific procedures and metrics. These participants requested that fairness experts pool knowledge and resources by domain for easy access. Additionally, practitioners in this study and \shortciteA{Veale2018} had concerns with the scalability of fairness analysis, which has been backed by data scientists in \shortciteA{Verma2018}. Practitioners interviewed by \shortciteA{Rakova2020} felt that metrics in academic-based research had vastly different objectives than metrics in industry, which required that industry researchers do the additional work to translate their work using insufficient metrics. To supplement this workload, practitioners felt that fairness research by academia did not fit smoothly into industry pipelines and required a significant amount of energy to fully implement \shortcite{Rakova2020}. 

In a study by \shortciteA{Richardson2021}, practitioners had the opportunity to interact with fairness tools and provide comments and feedback. Authors summarized key themes from practitioners regarding features and design considerations that would make these tools more applicable. The following lists contains some of the features requested for fairness toolkits:
\begin{itemize}
    \item Applicable to a diverse range of predictive tasks, model types, and data types \shortcite{Holstein}
    \item Can detect \& mitigate bias \shortcite{Holstein,Olteanu2019,Mehrabi}
    \item Can intervene at different stages of the ML/AI life cycle \shortcite{Bellamy2018,Holstein,Veale}
    \item Fairness and performance criteria agnostic  \shortcite{Corbett-Davies2018,Barocas2019,Verma2018}
    \item Diverse explanation types \shortcite{Ribeiro,Dodge2019,Arya,Binns2018}
    \item Provides recommendations for next steps \shortcite{Holstein}
    \item Well-supported with demos and tutorials \shortcite{Holstein}
\end{itemize}

The results from these studies collecting practitioner feedback supplement what many data scientists and social scientists have confirmed. In a study conducted by \shortciteA{Corbett-Davies2017}, when utilizing fairness metrics in a recidivism use case, results depicted that some metrics had significant trade-offs with public safety. This exemplified the importance of domain-specific guides and details when it comes to fairness procedures. In similar landscape summaries of algorithmic bias, \shortciteA{Rovatsos2019,Reisman2018} also discuss how concerns of algorithmic bias differ substantially by domain application, and yet most domains lack thorough and specific algorithmic bias guidance. Lastly, when studying the use of fair AI checklists, \shortciteA{Madaio2020,Cramer2018} emphasized the importance of aligning these checklists with team workflows. 

\subsubsection{Designing usable Fair AI} Concerns with applicability emphasize an over-arching importance of human-centered design procedures in the creation of fair AI tools. Efforts should be made to collect practitioner feedback and incorporate this feedback into the creation of fair AI \shortcite{Richardson2021,Holstein,Rakova2020}.

A major component of usability when it comes to fair AI tools is the integration of affordances that support AI fairness. According to \shortciteA{Robert2020}, these affordances include: transparency, explainability, voice, and visualization. In this work, they define transparency as ``making the underlying AI mechanics visible and known to the employee'', explainability as ``describing the AI’s decision/actions to the employee in human terms'', voice as ``providing employees with an opportunity to communicate and provide feedback to the AI'', and visualization as ``representing information to employees via images, diagrams, or animations'' \shortcite[p. 26]{Robert2020}. This work is not the only one of its kind to discuss the importance of such features in strengthening fairness tools.

Results from studies that interviewed or studied practitioners emphasized the importance of having the ability to interact with fairness toolkits \shortcite{Richardson2021,Holstein,Cramer2019}, using the tool to compare models and methods \shortcite{Richardson2021}, providing understandable visualizations \shortcite{Richardson2021,Veale2018,Law2020,Ribeiro,Dodge2019,Arya,Law,Veale}, and receiving feedback from these tools \shortcite{Ribeiro,Richardson2021}.  Solutions to aid users in understanding bias issues reside in interpretability and explainability research. \shortciteA{Hutchinson2018} emphasizes the importance of utilizing explanations in fairness pursuits. In a study by \shortciteA{Ribeiro}, participants had the opportunity to interact with LIME, the Local Interpretable Model-Agnostic Explanations toolkit. Results from this study depict the importance of explanations as a tool for users to decide whether or not they should trust a classifier or determine where and how to fix a classifier \shortcite{Ribeiro}. Without concrete explanations, users were either willfully ignorant or unable to rely on the classifier \shortcite{Ribeiro}. 

Furthermore, a study by \shortciteA{Dodge2019} found that different fairness problems were better explained with different types of explanations. When users were exposed to different types of fairness explanations, they exhibited very different opinions on the model. While some explanations were considered inherently less fair meaning users tended to view their models as biased, others enhanced participants' confidence in the classifier \shortcite{Dodge2019}. Authors in \shortciteA{Arya} did a similar study when introducing the AI Explainability 360 toolkit. They compared and contrasted different explainer types, including data, directly interpretable, local post-hoc, and global post-hoc explainers. They provide a taxonomy of explanations that provides guidance for which explainers are best for which users and future work that needs to be done \shortcite{Arya}.

\shortciteA{Law} ran a simulation with practitioners to depict how a plethora of visualizations could be used without overloading the participants. Depending on the use case, they proposed different techniques for organizing visualizations. One technique, which they referred to as a recommendation list, provides a summary of fairness results and was best for situations where there were many performance metrics. The other technique, visual cues, provided high comprehensiveness for a few metrics \shortcite{Law}. 

Despite the indubitable strengths of fairness toolkits, \shortciteA{Lakkaraju2019} and \shortciteA{Kaur} also gave some risks that came with the use of visualizations and explanations. \shortciteA{Lakkaraju2019} found that explanations for a black-box model could still be manipulated to hide issues of unfairness by simply omitting features that users considered problematic. By doing this, generated explanations increased user trust by nearly 10 fold \shortcite{Lakkaraju2019}. Furthermore, \shortciteA{Kaur} depicted the importance of training users before providing them with toolkits and visualizations that they are unfamiliar with. When users were not internalizing tutorials, they made incorrect assumptions about the data and the model, they could not successfully uncover issues with the dataset, and had low confidence when interacting with the tool. Nonetheless, they reported high trust in the fairness tools because it provided visualizations and it was publicly available \shortcite{Kaur}. These findings should be considered in the creation and the deployment of these toolkits.

\subsubsection{Avoid the over-reliance on Fair AI} As was depicted in the work of \shortciteA{Kaur}, many practitioners overly trusted toolkits despite not completely understanding the results of the toolkit. Of particular concern, when users were asked to select from a dropdown list the functionalities of the toolkit, most of them grossly overestimated what the toolkit could do. These results depicted a misalignment between practitioners’ understanding of the toolkits and the toolkits’ intended use \shortcite{Kaur}. Practitioners interviewed in \shortciteA{Veale2018} depicted similar hesitations about interacting with fairness tooling: they suspected they would over or under-rely on them.

\subsubsection{Communicating fairness to stakeholders} Across several interview studies where researchers collected practitioner feedback, practitioners requested help communicating fairness concerns to stakeholders \shortcite{Holstein,Veale2018,Law2020}. Some practitioners discussed how the objectives of stakeholders were different from their own and requested that resources be provided for communicating the fairness trade-off as it relates to those objectives \shortcite{Veale2018,Law2020}. They also requested that these resources be understandable by laymen, or business-oriented individuals, so that they can understand the cause of bias and the importance of fairness considerations \shortcite{Law2020,Garcia-Gathright2018}. One practitioner said that fairness toolkits should incorporate the assumption that there will be tech resistance and many stakeholders and practitioners will reject the output from fairness toolkits if its not effectively communicated \shortcite{Veale2018}. While prior discussions on different explanations for differing users could most definitely be useful in explaining fairness issues \shortcite{Arya}, fairness experts should consider creating guides for communicating fairness results.

\subsubsection{The burden of fairness} A common sentiment when it came to discussing or utilizing fairness tools was practitioners feeling overwhelmed. Whether users were interacting with a toolkit \shortcite{Law,Richardson2021} or a checklist \shortcite{Cramer2019}, many felt that either they were not qualified \shortcite{Holstein}, ethics was too difficult to operationalize \shortcite{Madaio2020}, ethics presented more questions than answers \shortcite{Binns2018a}, or the checklists and tools were too overwhelming \shortcite{Cramer2019,Richardson2021,Law}. While some of these concerns require support from institutions, many works discussed what fairness experts could do to relieve this burden. 

Since there exists so many avenues for bias and so many feasible solutions \shortcite{Mehrabi,Olteanu2019}, \shortciteA{Cramer2018} suggests that fairness experts create taxonomies of existing biases that can help practitioners recognize and counter these biases. To overcome the information overload that might accompany a visual toolkit \shortcite{Law}, \shortciteA{Law2020,Richardson2021} suggests that toolkits filter out biases on the users’ behalf so that alarming outcomes can stand out, but this solution presents selection biases of the fairness developers and important outcomes may differ in different contexts. Results from \shortciteA{Cramer2019} suggests that checklist be avoided as a ‘self-serve’ tool, since they were often found to be overwhelming for users. Instead, an interactive interface might be beneficial \shortcite{Cramer2019}. The goal for fairness practitioners should be to incorporate fairness into pre-existing workflows in such a way that disruption and chaos are minimized. With this objective in mind, fairness would be considered substantially less overwhelming for practitioners.

\subsubsection{Supplemental Resources for Practitioners} The major challenge presented to fairness experts is translating principles and ethics codes into actionable items that practitioners and institutions can implement \shortcite{Mittelstadt2019}. This includes providing guidance for users to detect and foresee bias \shortcite{Madaio2020,Law2020,Garcia-Gathright2018,Mehrabi,Cramer2018}, determine where bias might stem from \shortcite{Holstein,Garcia-Gathright2018}, determine how to respond to biases \shortcite{Holstein,Veale2018,Law2020,Garcia-Gathright2018}, and how to maintain fairness after deployment \shortcite{Veale2018}. Practitioners are requesting taxonomies of potential harms and biases \shortcite{Madaio2020,Cramer2018}, easy-to-digest summaries explaining biases \shortcite{Garcia-Gathright2018,Cramer2018}, and guidelines for best practices throughout the ML pipeline \shortcite{Holstein}. Users are also requesting a plethora of additional resources like community forums for fairness, domain-specific guides, and plenty of tutorials exemplifying how to incorporate fairness \shortcite{Holstein,Madaio2020,Richardson2021,Gray2019}. Fairness experts should do as much as possible to provide these resources for practitioners to support their use of fairness tools.

\subsection{Recommendations for Institutions}
While a considerable portion of recommendations were given to fairness researchers, organizations also have a large portion of responsibility for ensuring the success of fairness implementation. Several works provide critique and actionable steps for institutions and organizations to use to assist their practitioners in the implementation of fairness toolkits. 

\subsubsection{Prioritize Fairness} With the plethora of institutions who have created ethical principles or guidelines centered on fairness, accountability, and transparency in AI, there is no doubt that organizations have recognized the importance in prioritizing ethics \shortcite{Jobin2019}. Nonetheless, the application of fairness in practice is rare and most ethical codes are not put to action by engineers and practitioners. \shortciteA{Frankel1989} emphasizes that ethics codes that fail to put values into practice are political tools meant to manipulate the public. One of the easiest ways to promote a culture of bias awareness is to, at an organizational level, prioritize fairness as a global objective \shortcite{Garcia-Gathright2018,Stark2019}. Organizations can do this by considering fairness in their own hiring practices \shortcite{Garcia-Gathright2018}, prioritizing bias correction as they prioritize privacy and accessibility \shortcite{Madaio2020,Garcia-Gathright2018,Frankel1989}, providing practitioners with resources or teams that can provide them with actionable steps \shortcite{Madaio2020,Mittelstadt2019}, and changing organizational structure to embrace the inclusion of fairness considerations \shortcite{Cramer2019}. 

Practitioners interviewed in \shortciteA{Rakova2020} provided a plethora of recommendations for organizations intending to prioritize fairness. Organizations could provide reward systems to incentivize practitioners to continue educating themselves on ethical AI. Furthermore, they could allow practitioners to work more closely with marginalized communities to ensure their data is representative and their products are not harmful to them. Lastly, practitioners emphasized the importance of rejecting an AI system if it is found to perpetuate harms to individuals \shortcite{Rakova2020}.

\shortciteA{Reisman2018} recommended that organizations practice transparency with the public as a method for ensuring accountability and, therefore, fairness. They suggest that the public be given notice of any technology that might impact their lives \shortcite{Reisman2018}. Fairness research also emphasizes the importance of public transparency from institutions when it comes to the existence and the auditing of their AI \shortcite{Richardson2020}.

\subsubsection{Operationalize Fairness} One method to exhibit a prioritization of fairness is the operationalization of fairness. As they exist now, most published ethics codes are difficult to operationalize due to their abstract nature \shortcite{Madaio2020,Stark2019}. By operationalizing, organizations force themselves to consider how they want to enact fairness \shortcite{Robert2020,Frankel1989} and relieve the burden of ethical responsibility placed on the practitioner \shortcite{Stark2019} by making the big decision on things like: what type of demographics, biases, and stakeholders they intend to consider in their efforts \shortcite{Cramer2018}. This will require discussions by top-players within organizations in deciding how the prioritization of fairness may displace other priorities, and it requires the consultation of legal-consultants who can ensure new policies satisfy government regulations \shortcite{Veale2018,Mittelstadt2019}. Furthermore, leaders must make the decision of how much bias is permissible and be prepared to handle the consequences of these decisions \shortcite{Barocas2018,Barocas2019}.

\subsubsection{Avoid dividing practitioner loyalty} According to recent research, many practitioners feel divided by the desire to create ethically-sound products and the obligation to their role as employees \shortcite{Stark2019,Mittelstadt2019}. Some felt as though conversations of ethics were taboo and could impact their goals of career advancement \shortcite{Madaio2020,Rakova2020}. Many felt like it was an individual battle, where they were the sole advocate ‘battling’ for fairness \shortcite{Holstein,Madaio2020}. 

When studying ethics in engineering, it becomes obvious that these concerns are nothing new to the field of engineering. Works by \shortciteA{Frankel1989} and \shortciteA{Davis1991} discuss the ethical dilemma that has impacted engineers for decades. When studying the Code of Ethics taught to engineers, it is clear that engineers should prioritize ethical obligation over institutional obligations and call out those who do not \shortcite{Davis1991}. Nonetheless, this rhetoric is rarely implemented and those who attempt to follow codes are often referred to as “whistle-blowers”. Institutions can prevent this by  creating clear guidelines for what practitioners should do when they encounter these issues and rewarding individuals who recognize biases and follow the appropriate steps to fixing them \shortcite{Frankel1989}.

\subsubsection{Create Organizational structure around ethics} When interviewed, practitioners stated concerns based on the lack of fairness infrastructure within organizations. Many practitioners stated that most fairness mitigation was done independently by practitioners and often their efforts went uncompensated \shortcite{Madaio2020,Rakova2020}. Others said they were unsure of who to share fairness issues with or if it was even their responsibility to do so \shortcite{Veale2018,Rakova2020,Stark2019}. Lastly, some practitioners had concerns with how to handle fairness issues when different parts of an AI product are owned and/or handled by different teams \shortcite{Cramer2018}. 

These comments suggest a dire need for organizations to establish clear delegations of tasks around fairness. Some practitioners even recommended that fairness teams and internal review boards be created to be used as a resource for practitioners and as auditors of AI technologies \shortcite{Rakova2020,Robert2020}. With these additional support teams, there should be a formal path for communication that dictates clearly to practitioners the hierarchy for communicating relevant issues \shortcite{Veale2018}. If issues arise as it relates to fairness, organizations should have prepared plans for how these issues will be dealt with \shortcite{Rakova2020}. Organizations should consider the use of internal and external investigation committees, especially if handling sensitive protected attributes \shortcite{Rakova2020,Veale}. It is critical that these teams be given open access to the data, the models, and the procedures in order to optimize transparency and reproducibility \shortcite{Haibe-Kains2020}. Additionally, education infrastructure, focusing on educating practitioners on recognizing and combating algorithmic bias, should be made available to practitioners and product managers to ensure a baseline of understanding amongst practitioners \shortcite{Rakova2020}.

The prioritization of fairness by organizations is critical for their practitioners, their global image, and the consumers of their products and services. Organizations can prioritize fairness by operationalizing it, defining an explicit infrastructure for it, and ensuring protection for employees interested in implementing it.

\subsection{Recommendations for ML Practitioners}
Despite the predominant portion of work that needs to be done by fairness experts and practitioners, there are still a few recommendations in literature for ML practitioners. In order to ensure the success of fairness, effort must also be made by practitioners to educate themselves and properly implement fairness when it is made available to them. This section briefly provides recommendations for how practitioners can correctly implement fairness into their pipelines.

\subsubsection{Implement fairness throughout the pipeline} As stated by \shortciteA{Stark2019}, ethics cannot be considered a final checkpoint but should be considered from the very start of a project and should be reflected within every stage of the process. Currently, most practitioners engage with fairness ad-hocly or as a final performance check \shortcite{Holstein,Madaio2020,Rakova2020}. But the literature suggests that fairness would be much easier to implement if it is considered from the idea formulation stage of the project \shortcite{Cramer2018,Cramer2019,Friedman2001}. The task becomes much more difficult when a model has been deployed and running for a while because those unintended biases will have become recursive and, therefore, much more difficult to manage \shortcite{Cramer2018}. Furthermore, practitioners should make an effort in including a wide variety of experts and lay-persons into the design, implementation, and evaluation of ML \shortcite{Wagstaff2012}. Practitioners should make sure that assessing and addressing bias becomes a normal procedure throughout the pipeline of their projects.

\subsubsection{Iteratively implement fairness} While fairness should be considered throughout the process, sometimes it may be overwhelming to immediately counteract every instance of unfairness found. When \shortciteA{Cramer2018} allowed practitioners to engage with fairness checklists, they noticed concerns practitioners had with how these checklists might conflict with the rapid delivery development procedures that existed within their organizations. \shortciteA{Cramer2018} concluded that addressing algorithmic bias would need to be done in short-term narrow steps with plans for iterative improvement of fairness issues. \shortciteA{Cramer2019} also suggested that such a method could reduce the overload felt by practitioners. Therefore, practitioners should iteratively implement small fixes to fairness issues between deployments and not try to detect and handle every issue immediately. 
	
\section{Related works}
\label{relatedworks}
While this work is the first of its kind to provide a comprehensive review of the solution space of fairness AI, there are a few related reviews. Section 2’s summative breakdown of biases that exists in machine learning are similar to that of \shortcite{Friedman1996,Mehrabi,Olteanu2019,Hutchinson2018,Hellstrom2020}. \shortciteA{Mehrabi} provides a comprehensive list of 23 types of biases, which are not organized in any format. \shortciteA{Friedman1996} and \shortciteA{Olteanu2019} both provided a hierarchy of biases, creating their own structure for how these biases should be organized. While \shortciteA{Olteanu2019} summarizes the different biases that are associated with social data, \shortciteA{Friedman1996} defines a new set of biases around data in machine learning. This work adopts part of \shortciteA{Friedman1996}’s hierarchical structure and utilizes thematic analysis to categorize different biases introduced across a comprehensive list of works. \shortciteA{Hutchinson2018} discusses the history of fairness, including the differing notions of fairness and gaps in fairness work. Lastly, \shortciteA{Hellstrom2020} creates their own taxonomy of bias, discusses the relation between forms of bias, and provides examples to support differing definitions. Their taxonomy, however, is depicted based on chronological influences for bias.

While, to our knowledge, there are no reviews that exists that are similar to our Section 3, there are some works that do a user-focused comparative analysis of different toolkits. \shortciteA{Richardson2021} is a paper that allows practitioners to engage with two toolkits, UChicago’s Aequitas \shortcite{Saleiro2018} and Google’s What-if \shortcite{Wexler2019} and collects their feedback. \shortciteA{Seng} is a similar work that compares six different toolkits: Scikit’s fairness tool \shortcite{Johnson2020}, IBM Fairness 360 \shortcite{Bellamy2018}, UChicago’s Aequitas \shortcite{Saleiro2018}, Google’s What-if \shortcite{Wexler2019}, PyMetrics Audit-AI, and Microsoft’s Fairlearn \shortcite{Bird2020}. To our knowledge, there is no other work that reviews both fairness toolkits and checklists.

Furthermore, our contributions from Section 4 are entirely novel. While \shortciteA{Richardson2021} does a brief literature review of related works that take practitioner feedback into consideration, this paper is the first of its kind to combine recommendations from practitioners, fairness experts, and institutions alike and presents a comprehensive review of necessary work for fairness research.

\section{Conclusion}
\label{conclusion}
There exists a seemingly endless list of potential influences of bias when it comes to machine learning algorithms. In order to respond to the prolific nature of bias, fairness researchers have produced a wide variety of tools that practitioners can use, mainly in the form of software toolkits and checklists. Companies, academic institutions, and independent researchers alike have created their own versions of these resources, each with unique features that those developers thought were critical. Nonetheless, there still exists a disconnect between the fairness developers and the practitioners that prevents the use of these resources in practice. The literature suggests that fairness in practice can be normalized at the effort of fairness developers, institutions who create AI, and practitioners who build AI.

This review provides a summary of algorithmic bias issues that have arisen in literature, highlights some unique contributions in the fairness solution space, and summarizes recommendations for the normalization of fairness practices. To our knowledge, there exists no other work that provides such a comprehensive review of the problem and solution space of fairness research. The literature suggests that there is still much work to be done and this review highlights those major needs. 

\acks{The authors wish to thank Hans-Martin Adorf, Don Rosenthal, 
Richard Franier, Peter Cheeseman and Monte Zweben for their assistance
and advice.  We also thank Ron Musick and our anonymous reviewers for
their comments.  The Space Telescope Science Institute is operated by
the Association of Universities for Research in Astronomy for NASA.
}

\vskip 0.2in
\bibliography{sample}
\bibliographystyle{theapa}

\end{document}